\documentclass[10pt,twocolumn,letterpaper]{article}

\usepackage{cvpr}              
\usepackage{stfloats}
\definecolor{cvprblue}{rgb}{0.21,0.49,0.74}
\usepackage[pagebackref,breaklinks,colorlinks,allcolors=cvprblue]{hyperref}
\usepackage{float}
\usepackage{subcaption} 

\def\paperID{57} 
\def\confName{CVPR}
\def\confYear{2025}

\title{Unsupervised Pelage Pattern Unwrapping for Animal Re-identification}

\author{Aleksandr Algasov*\textsuperscript{\textdagger}, Ekaterina Nepovinnykh*\textsuperscript{\textdaggerdbl}, Fedor Zolotarev*, Tuomas Eerola*, Heikki K\"{a}lvi\"{a}inen*\textsuperscript{\textdagger}, \\ Pavel Zem\v{c}\'{i}k\textsuperscript{\textdagger}*, Charles V. Stewart\textsuperscript{\textdaggerdbl}\\
\\
		*Computer Vision and Pattern Recognition Laboratory (CVPRL)\\
	 Lappeenranta-Lahti University of Technology LUT, Lappeenranta, Finland\\
		{\tt\small firstname.lastname@lut.fi}\\
    \textsuperscript{\textdagger}Faculty of Information Technology \\
    Brno University of Technology (BUT), Brno, Czech Republic \\
    {\tt\small zemcik@fit.vut.cz}\\       
    \textsuperscript{\textdaggerdbl}Department of Computer Science \\
    Rensselaer Polytechnic Institute (RPI), Troy, NY, USA \\
    {\tt\small stewart@rpi.edu}
}

\begin{document}
\maketitle
\begin{abstract} 
Existing individual re-identification methods often struggle with the deformable nature of animal fur or skin patterns which undergo geometric distortions due to body movement and posture changes. In this paper, we propose a geometry-aware texture mapping approach that unwarps pelage patterns, the unique markings found on an animal’s skin or fur,  into a canonical UV space, enabling more robust feature matching. Our method uses surface normal estimation to guide the unwrapping process while preserving the geometric consistency between the 3D surface and the 2D texture space. We focus on two challenging species: Saimaa ringed seals (Pusa hispida saimensis) and leopards (Panthera pardus). Both species have distinctive yet highly deformable fur patterns. By integrating our pattern-preserving UV mapping with existing re-identification techniques, we demonstrate improved accuracy across diverse poses and viewing angles. Our framework does not require ground truth UV annotations and can be trained in a self-supervised manner. Experiments on seal and leopard datasets show up to a 5.4\% improvement in re-identification accuracy.
\end{abstract}
    
\begin{figure*}[bp]
    \centering
    \includegraphics[width=\linewidth]{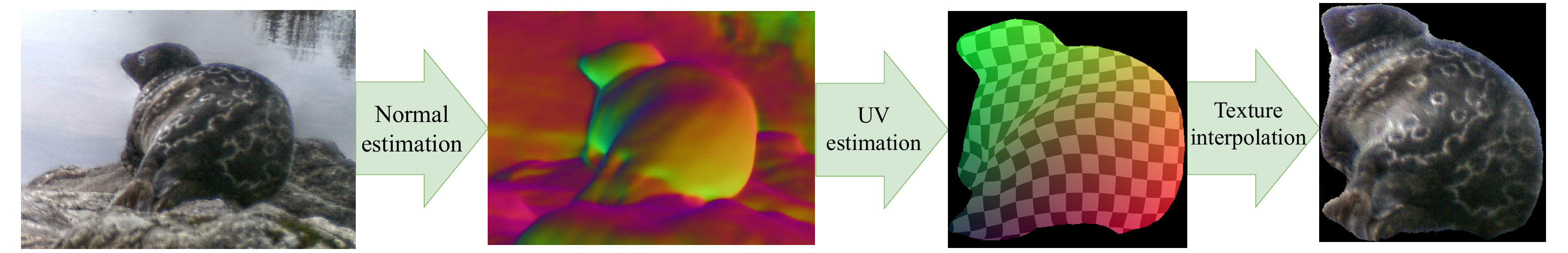}
    \caption{Overview of the pelage pattern unwrapping pipeline.}
    \label{fig:pipeline}
\end{figure*}
\section{Introduction}

Monitoring wildlife populations is part of many conservation efforts and ecological research. Individual animal identification enables researchers to track movement patterns, estimate population sizes, and study behavioral dynamics. Computer vision-based re-identification offers a non-invasive methods of identifying individuals based on their natural appearance patterns such as 
fur and skin~\cite{nepovinnykh2024alfreid, nepovinnykh2024norppa, otarashvili2024multispecies, vcermak2024wildlifedatasets}, feather patterns~\cite{deeplearningbirds}, ear ~\cite{de2022reidentifying, kulits2021elephantbook} and fin shapes~\cite{hughes2017automated, blount2022flukebook}, as well as facial features~\cite{deb2018face, shinoda2024petface}.

Many species exhibit unique visual markings that can be used for identification. For instance, zebras have unique stripe patterns, tigers have distinctive fur patterns, and Saimaa ringed seals have characteristic ring-shaped patterns on their fur. However, depending on the species the pattern can have deformable nature causing parts of these visual traits may become hidden or distorted due to movement or body posture. For example, marine mammals like seals and whales have dynamic and less predictable body shapes in images. This poses a challenge for end-to-end deep learning methods that rely heavily on rigid pattern alignment. At the same time, the visual differences in the patterns that distinguish two individuals can be very subtle. This means that the image features used for re-identification must be both invariant to possible deformations and sensitive to fine differences in the shapes present in the patterns. Such features are inherently difficult to learn.

UV (unwrapped visual texture) map estimation has been extensively studied in the context of human-centric vision, where it serves to map pixels in 2D images to canonical 3D surfaces for tasks like pose analysis, garment editing, or activity recognition. Classical approaches like DensePose~\cite{guler2018densepose} and continuous surface embeddings (CSE)~\cite{neverova2020continuous} estimate these maps to enable dense correspondence between image and body surface, and more recent methods like PC-CSE~\cite{perez2025human} integrate pose constraints to produce anatomically coherent maps. Similarly, in the garment domain, normal-guided UV estimation methods utilize 3D surface normals and isometry constraints to produce temporally consistent UV maps across frames without requiring full 3D reconstruction~\cite{jafarian2023normal}. These techniques allow for fine-grained alignment of appearance patterns despite complex deformations and are especially effective for dynamic and non-rigid objects, such as clothing. 

In this paper, we draw inspiration from human garment re-texturing techniques to address the challenges of animal pattern deformation. We propose a novel approach that estimates a geometry-aware UV mapping to unwrap animal pelage patterns into a canonical texture space. The overview of the proposed pipeline is presented in Fig.~\ref{fig:pipeline}. Our main idea is that by mapping the curved 3D surface of an animal to a 2D texture space while preserving isometry, we can obtain a representation that is invariant to body posture and deformation. Although UV estimation has not yet been applied to animal re-identification, its ability to disentangle appearance from pose makes it a promising direction for species with highly deformable bodies, such as marine mammals.

Our approach consists of three main components: (1) surface normal estimation from animal images to capture the underlying 3D geometry, (2) geometry-aware UV mapping that preserves isometry between the 3D surface and 2D texture space, and (3) pattern matching in the unwrapped texture space. The unwrapped patterns allow for more accurate comparison across different images of the same individual, regardless of body pose and deformation.

Our main contributions are: (1) We propose the first UV mapping-based framework for animal re-identification to address pattern deformation in highly flexible species.
(2) We introduce a self-supervised method for estimating geometry-aware UV coordinates from animal images without requiring ground truth annotations. (3) We demonstrate the effectiveness of our approach on two challenging species: Saimaa ringed seals and leopards, showing improved re-identification accuracy under significant pose variation on both species.

\section{Method}

Our goal is to develop a robust animal re-identification system that can handle significant body deformations by unwrapping pelage patterns into a canonical texture space. The key insight is that by mapping the curved 3D surface of an animal to a 2D texture space while preserving isometry, we can obtain a representation that is invariant to body posture and deformation.

\subsection{Overview}

Fig.~\ref{fig:pipeline} illustrates the overview of our approach. Given an input image of an animal, we first estimate surface normals to capture the underlying 3D geometry. We then predict a UV mapping that preserves isometry with respect to the estimated 3D surface. This UV mapping allows us to unwrap the pelage pattern into a canonical texture space. Finally, we apply existing re-identification techniques on the unwrapped patterns to match individuals across different images.

\subsection{Instance Segmentation}

The instance segmentation is needed to isolate animal and its surface normals in the image, in order to improve UV-mapping algorithm performance, as background information is irrelevant and can add confusion to UV-estimating process. The segmentation is done using SEEM~\cite{seem}, which is promptable and interactive model for segmenting, capable of generating segmentation masks for animals, which exclude most of the irrelevant information.

\subsection{Surface Normal Estimation}

To capture the 3D geometry of the animal surface, we estimate per-pixel surface normals from the input image. We use Metric3Dv2~\cite{Hu_2024_metric3dv2} to do the estimation without any fine-tuning or modifications to the original method. It is a state-of-the-art method for monocular depth and normal estimation from natural images. It should be noted, that unlike with the rest of the method stages, to estimate normals we use original images of the animals with the background. This choice is motivated by the observation that Metric3Dv2, like many monocular depth and normal estimation models, is trained on natural images that include background context. Removing the background during inference can disrupt the spatial cues these models rely on, leading to degraded or unstable predictions, especially near object boundaries. Using the original image helps preserve the environmental context necessary for reliable normal estimation over the animal's surface.

\subsection{Geometry-Aware UV Mapping}

The method of Jafarian et al.~\cite{jafarian2023normal}  is used to predict a UV mapping that preserves isometry between the 3D animal surface and the 2D texture space. The mapping is defined as 
\begin{equation}
    g: (x, y) \mapsto (u, v),
\end{equation}
where $(x, y)$ are coordinates of pixels in the original image and $(u, v)$ are UV coordinates.
The method leverages 3D surface normals to guide the unwrapping process by formulating partial differential equations that encode the isometry constraint. The mapping is modeled as a multi-layer perceptron and the system of equations is solved by minimizing a special loss function. We modified the method by further adding partial derivatives clipping to better handle singularities (e.g. sharp changes in gradient direction). 

\subsection{Texture Unwrapping}

To generate the unwrapped texture, we first collect all available pixels from the original image along with their corresponding UV coordinates produced by the mapping \( g \). To assign a UV coordinate to each pixel in the canonical texture map, we apply Delaunay triangulation to a set of known UV coordinates. This creates a mesh that defines how the surface is subdivided in the UV space. For each pixel in the target unwrapped texture, we identify the triangle it falls into and interpolate its color using barycentric coordinates derived from the colors of the triangle’s vertices.

\section{Experiments}

\subsection{Implementation Details}

Normal estimations were performed on the original unsegmented images using \verb|metric3d_vit_giant2| variant of Metric3Dv2. For segmentation task SEEM model with Focal-L backbone was used, prompting the name of the species to guide the method. The neural network for UV mapping estimation consists of Fourier positional encoding layer and six dense layers with ReLU activation function. The architecture follows the same structure as defined in~\cite{jafarian2023normal}, with the only modification of reduced number of layers. The training was performed over 100 epochs of pretraining and then 300 epochs of training, both with cosine annealing learning rate scheduler, starting with learning rate of $10^{-5}$.

\begin{table*}[ht]
\centering
\caption{Evaluation results for various descriptors on the SealID dataset: Top-N and mAP are expressed in percentages. “Failures” and “Improvements” indicate the number of Rank-1 searches that, respectively, failed or succeeded with unwrapped images compared to the original images. The best mAP and Top-N metric values are highlighted in bold.}
\label{tab:evaluation_results}
\begin{tabular}{lccccccc}
\toprule
Name & mAP & Top-1 & Top-3 & Top-5 & Top-10 & Failures & Improvements \\
\midrule
ALIKED - original       & 18.5 & 39.2 & 47.3 & 50.6 & 52.7 &       &       \\
ALIKED - unwrapped      & 19.8 & 41.6 & 49.0 & 51.6 & 56.5 & 18    & 25    \\
\addlinespace
DogHardNet - original   & 20.4 & 45.4 & 51.5 & 52.7 & 55.4 &       &       \\
DogHardNet - unwrapped  & 21.9 & \textbf{48.8} & \textbf{54.6} & \textbf{56.3} & \textbf{59.7} & 21    & 31    \\
\addlinespace
SuperPoint - original   & 19.2 & 43.3 & 49.0 & 50.4 & 52.9 &       &       \\
SuperPoint - unwrapped  & 22.2 & 48.0 & 53.5 & \textbf{56.3} & 58.4 & 20    & 36    \\
\addlinespace
DISK - original         & 12.7 & 27.9 & 35.6 & 38.3 & 42.7 &       &       \\
DISK - unwrapped        & 15.4 & 33.3 & 41.8 & 44.4 & 46.9 & 20    & 44    \\
\addlinespace
ALFRE-ID - original      & 28.7          & 45.0 & 51.0 & 52.7 & 56.7 &  & \\
ALFRE-ID - unwrapped     & \textbf{30.0} & 47.6 & 53.3 & 54.6 & 57.3 & 21 & 31 \\
\bottomrule
\end{tabular}
\end{table*}

\begin{table*}[ht]
    \centering
    \caption{Evaluation results for various descriptors on the Leopard dataset: Top-N and mAP are expressed in percentages. “Failures” and “Improvements” indicate the number of Rank-1 searches that, respectively, failed or succeeded with unwrapped images compared to the original images. The best mAP and Top-N metric values are highlighted in bold.}
    \label{tab:leopard-performance}
    \begin{tabular}{lccccccc}
    \toprule
    Name      & mAP   & Top-1  & Top-3  & Top-5  & Top-10 & Failures & Improvements \\
    \midrule
    ALIKE-D - original        & 52.3  & 82.2   & 86.0   & 87.1   & 87.7  & & \\
    ALIKE-D - unwrapped       & 44.2  & 70.8   & 76.5   & 77.8   & 79.5  & 117 & 10 \\
    \addlinespace
    DogHardNet - original     & 52.3  & 85.8   & 87.5   & 88.5   & 88.8  & & \\
    DogHardNet - unwrapped    & \textbf{53.7}  & \textbf{87.6}   & \textbf{89.1}   & \textbf{89.9}   & \textbf{90.1}  & 17 & 34 \\
    \addlinespace
    SuperPoint - original     & 43.8  & 72.9   & 78.7   & 80.1   & 81.2  & & \\
    SuperPoint - unwrapped    & 42.6  & 72.1   & 77.6   & 79.5   & 80.8  & 56 & 48 \\
    \addlinespace
    DISK - original           & 49.2  & 79.1   & 83.0   & 84.8   & 85.7  & & \\
    DISK - unwrapped          & 46.5  & 78.0   & 81.7   & 82.8   & 83.4  & 52 & 42 \\
    \addlinespace
    ALFRE-ID - original        & 46.8  & 66.5   & 71.3   & 73.5   & 76.1  & & \\
    ALFRE-ID - unwrapped       & 44.8  & 65.7   & 70.5   & 72.8   & 75.1  & 41 & 33 \\
    
    \bottomrule
\end{tabular}

\end{table*}

\begin{figure*}[th]
    \centering
    \includegraphics[width=0.97\textwidth]{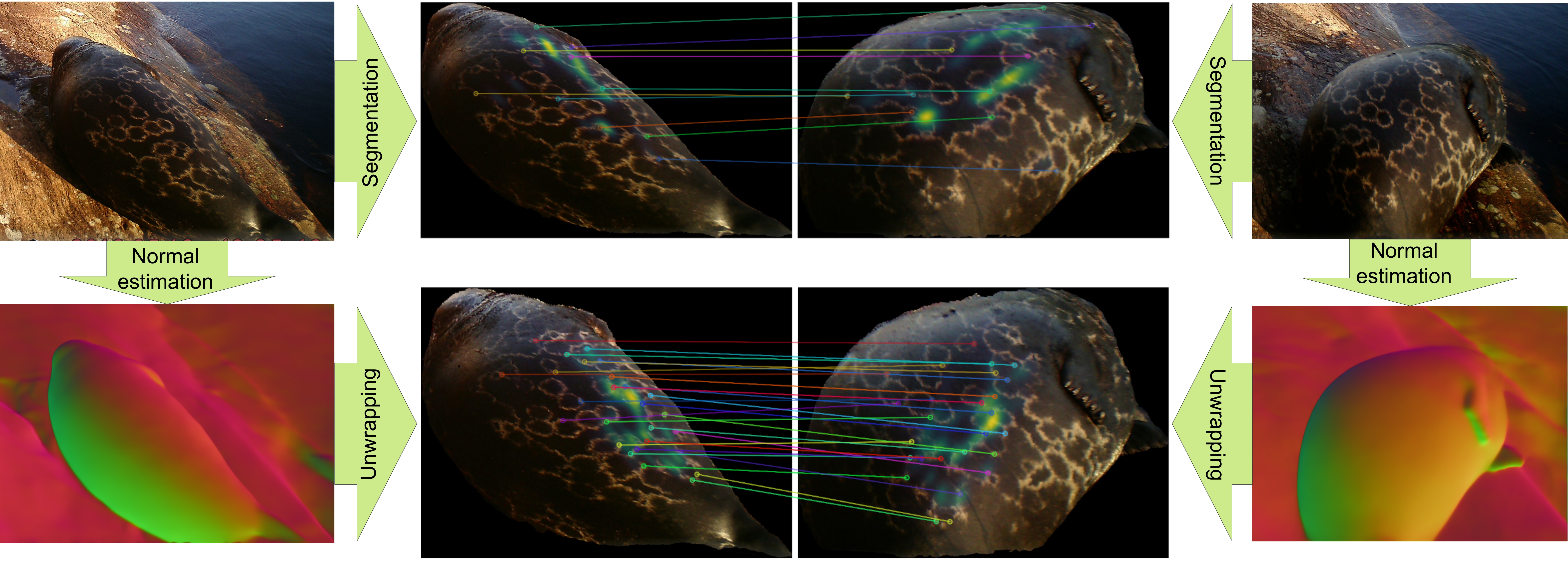}
    \caption{A visualization of the proposed method and how texture unwrapping can imporve feature matching. The top row shows LightGlue matches on the segmented images. On the bottom, the textures are unwrapped based on the estimated normals. Density of extracted features is highlighted in green, opacity of connecting lines corresponds to the score of the match.  }
    \label{fig:matching_example_seal}
\end{figure*}

\begin{figure*}[th]
    \centering
    \begin{subfigure}{0.8\textwidth}
    \centering
    \includegraphics[width=\textwidth]{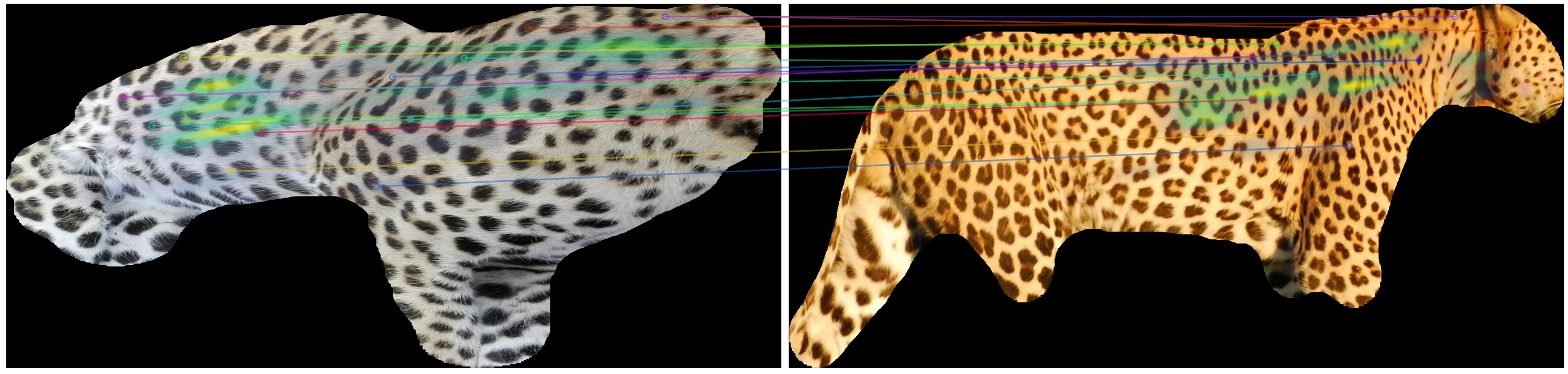}
    \caption{Matches found by LightGlue on the segmented images of a leopard.}
    \label{subfig:match1}
    \end{subfigure}
    \begin{subfigure}{0.8\textwidth}
    \centering
    \includegraphics[width=\textwidth]{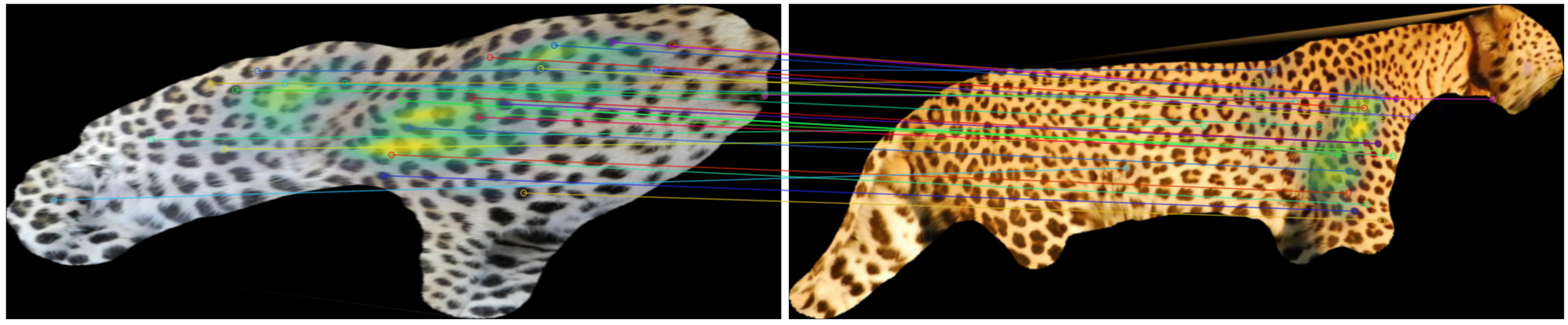}
    \caption{Matches found by LightGlue on the unwrapped textures of a leopard.}
    \label{subfig:match2}
    \end{subfigure}
    \caption{A visualization of matched features. Density of extracted features is highlighted in green, opacity of connecting lines corresponds to the score of the match. Note how the unwrapped texture removed the crease around the front paw and produced stronger matches in that region. }
    \label{fig:matching_example_leopard}
\end{figure*}

\subsection{Design}
We evaluated our method on test splits of a dataset of Saimaa ringed seals, which is a subset of the SealID dataset~\cite{nepovinnykh2022sealid} (480 annotations with 146 individual animals), and the African leopard dataset (938 annotations with 256 individual animals), provided by the African Carnivore Wildbook (ACW).
The evaluation was done using a leave-one-out approach, in which one image from a set was designated as a query while the remaining images formed the database. This procedure was repeated for every image in the set. The matching of individuals (re-identification) was done using LightGlue~\cite{lindenberger2023lightglue} and ALFRE-ID~\cite{nepovinnykh2024alfreid}. LightGlue was used with four different local feature descriptors SuperPoint~\cite{detone2018superpoint}, DISK~\cite{tyszkiewicz2020disk}, ALIKED~\cite{zhao2023alikedlighterkeypointdescriptor}, and DoGHardNet~\cite{HardNet2017}. The matches were obtained from similarity score between a query and a database images, based on the area under the curve of match scores provided by LightGlue. ALFRE-ID was utilized in the same manner as in original paper, including additional steps of tonemapping and pattern extraction for the SealID dataset.

The method was applied to the test split of the dataset, since fine-tuning is not required for LightGlue to do the matching. To establish a baseline, the evaluation was performed on segmented images of animals without texture unwrapping. The same images were then used to run the method and create an unwrapped variant of the test set. 
The evaluation process was done on both variants of unwrapped test sets measuring Top-k, mAP metrics, and improvements (re-identification failed on original images but succeeded on unwrapped images) and failures (re-identification succeeded on original images but failed on unwrapped images) in re-identification.

\section{Results and Discussion}

The results for the SealID dataset are presented in Table~\ref{tab:evaluation_results}. Unwrapping consistently boosts performance for all descriptors, as well as for ALFRE-ID. Notably, DISK sees its mAP improve from 12.7\% to 15.4\% and its Top-1 accuracy gains 5.4\%. The gain in accuracy and mAP for the best performing methods, LightGlue+DogHardNet and ALFRE-ID, is about 2-3\%. The overall improvement supports the idea that minimizing the natural deformations improves the recognition of pelage pattern in otherwise warped areas. 

Visualization of matches obtained with and without unwrapping is presented in Fig.~\ref{fig:matching_example_seal}. All visualizations use the LightGlue + DogHardNet to generate matches. The proposed method is especially helpful in removing the distortion due to the skin stretching and folding, as well as the natural distortion due to the grazing angles of view. After unwrapping, LightGlue produced not only a larger number of matches, but the scores of those matches were also much higher when compared to the original images, making it more confident in its overall score. 

The results on the leopard dataset are more varied, as demonstrated in Table~\ref{tab:leopard-performance}. Out of all methods, only LightGlue + DogHardNet showed an improvement, while all other methods, including ALFRE-ID, have degraded performance on the unwrapped textures. However, it should be noted that in overall the best re-identification performance was obtained on unwrapped images (LightGlue + DogHardNet). For most methods, the accuracy did not decrease by more than a 1\%, except for LightGlue + ALIKE-D, where the difference is greater than 10\%. The improvement for LightGlue + DogHardNet is around 1-2\% for all metrics. This discrepancy showcases the vast difference in descriptor performance, highlighting the importance of using the right descriptors for the task. An overall underwhelming performance of the method on the leopard dataset showcases the main weakness of the proposed approach, as it is highly dependent on the accuracy of the normal estimation, which can often fail in low-light and obstructed environments, as is often a case in camera trap datasets. 

A visualization of the matches between an example pair of images is presented in Fig.~\ref{fig:matching_example_leopard}. It can be seen that the method successfully dealt with naturally occurring distortions due to the variation in the pose of an animal and produced more matches in the otherwise challenging parts of the texture.

\section{Conclusion}

In this paper, we showed a potential improvement for image-based animal re-identification that leverages geometry-aware texture mapping to unwrap pelage patterns into a canonical texture space. By preserving the isometry between the 3D surface and the 2D texture space, the method creates representations that are robust to body deformations and grazing viewing angles. Experiments on Saimaa ringed seals demonstrate improvements in matching unwrapped images over original images. Results on the leopard dataset indicate the potential for improvement, but are dependent on the quality of the descriptors and normal estimation.

Future work includes applying the method to other species with distinctive patterns. We also plan to improve the unwrapping algorithm by aligning the UV texture according to a species-specific orientation, potentially removing the rotation and translation uncertainty. Additionally, we will look at whether supervised end-to-end methods can benefit from unwrapped textures.

\section*{Acknowledgments}

The authors would like to thank the Finnish Cultural Foundation for funding the research. In addition, the authors would like to thank the Department of Environmental and Biological Sciences at the University of Eastern Finland (UEF) for providing data on Saimaa ringed seals, the African Carnivore Wildbook (ACW) for providing African leopard data. 
{
    \small
    \bibliographystyle{ieeenat_fullname}
    \bibliography{main}
}


\end{document}